\documentclass[runningheads]{llncs}

\usepackage{eccv}

\usepackage{eccvabbrv}

\usepackage{graphicx}
\usepackage{booktabs}

\usepackage[accsupp]{axessibility}  %

\usepackage[pagebackref,breaklinks,colorlinks,citecolor=eccvblue]{hyperref}

\usepackage{orcidlink}

\newcommand{\todo}[1]{{\color{red}#1}}
\newcommand{\TODO}[1]{\textbf{\color{red}[TODO: #1]}}

\usepackage{adjustbox}
\usepackage{graphicx}
\usepackage{scalerel}
\usepackage{array}
\usepackage{caption}
\usepackage{subcaption}
\usepackage{printlen}
\usepackage{float}
\usepackage{multirow}
\usepackage{wrapfig}
\usepackage{booktabs}
\usepackage{pifont}
\usepackage{makecell}
\usepackage{nicefrac}
\usepackage{epigraph}
\usepackage{wrapfig}
\usepackage{url}
\usepackage{stmaryrd}
\usepackage{verbatim}

\usepackage{algorithm}
\usepackage{algpseudocode}

\usepackage{pgf}
\usepackage{pgfplots}
\usepackage{ifthen}
\usepackage{soul}
\usepackage{tabularx}
\usepackage{tikz}

\usepackage[capitalize,noabbrev]{cleveref}

\usepackage{microtype}
\usetikzlibrary{angles,quotes,3d,math,arrows.meta,calc,positioning,fit,backgrounds,decorations.pathreplacing,calligraphy,shapes,shapes.multipart}
\usepackage[table]{xcolor}

\renewcommand{\todo}[1]{{\color{red}#1}}
\renewcommand{\TODO}[1]{\textbf{\color{red}[TODO: #1]}}
\newcommand{\stefan}[1]{\textbf{{\color{ourgreen}[Stefan: #1]}}}
\newcommand{\nick}[1]{\textbf{{\color{ourturquoise}[Nick: #1]}}}
\newcommand{\kolja}[1]{\textbf{{\color{ourblue}[Kolja: #1]}}}

\newcommand{\omitforsubmission}[1]{{\color{ourpurple}#1}}
\newcommand{\remove}[1]{{\color{ourorange}#1}}
\newcommand{\draft}[1]{{\color{gray}#1}}
\newcommand{\compactversion}[1]{#1}

\newcommand{\setlearner}{set learner}

\renewcommand{\TODO}[1]{}
\renewcommand{\todo}[1]{#1}
\renewcommand{\nick}[1]{}
\renewcommand{\stefan}[1]{}
\renewcommand{\kolja}[1]{}
\renewcommand{\compactversion}[1]{}
\renewcommand{\omitforsubmission}[1]{}
\renewcommand{\draft}[1]{}
\renewcommand{\remove}[1]{}

\crefname{appsec}{Supp.\ Sec.}{Supp.\ Secs.}
\Crefname{appsec}{Supp.\ Section}{Supp.\ Sections}
\crefname{appfig}{Supp.\ Fig.}{Supp.\ Figs.}
\Crefname{appfig}{Supp.\ Figure}{Supp.\ Figures}
\crefname{apptab}{Supp.\ Tab.}{Supp.\ Tabs.}
\Crefname{apptab}{Supp.\ Table}{Supp.\ Tables}
\crefname{appeq}{Supp.\ Eq.}{Supp.\ Eqs.}
\Crefname{appeq}{Supp.\ Equation}{Supp.\ Equations}

\definecolor{ourgreen}{RGB}{46, 204, 113}
\definecolor{ourgreenborder}{RGB}{39, 174, 96}
\definecolor{ourblue}{RGB}{52, 152, 219}
\definecolor{ourblueborder}{RGB}{41, 128, 185}
\definecolor{ourorange}{RGB}{230, 126, 34}
\definecolor{ourorangeborder}{RGB}{211, 84, 0}
\definecolor{ourred}{RGB}{231, 76, 60}
\definecolor{ourredborder}{RGB}{192, 57, 43}
\definecolor{ouryellow}{RGB}{241, 196, 15}
\definecolor{ouryellowborder}{RGB}{243, 156, 18}
\definecolor{ourpurple}{RGB}{155, 89, 182}
\definecolor{ourpurpleborder}{RGB}{142, 68, 173}
\definecolor{ourturquoise}{RGB}{26, 188, 156}
\definecolor{ourturquoiseborder}{RGB}{22, 160, 133}
\definecolor{ourturquoise}{RGB}{26, 188, 156}
\definecolor{ourturquoiseborder}{RGB}{22, 160, 133}
\definecolor{ourwhite}{RGB}{236, 240, 241}
\definecolor{ourwhiteborder}{RGB}{189, 195, 199}
\definecolor{ourgray}{RGB}{149, 165, 166}
\definecolor{ourgrayborder}{RGB}{127, 140, 141}

\definecolor{ourwhite2}{RGB}{246, 247, 248}

\definecolor{ourhighlightcolor}{RGB}{46, 204, 113}

\newcolumntype{H}{>{\setbox0=\hbox\bgroup}c<{\egroup}@{}}

\newcolumntype{C}[1]{>{\centering\arraybackslash}m{#1}}

\newcommand{\tikzstylenodedistance}{4mm}
\newcommand{\tikzstyleinnersep}{2mm}
\newcommand{\tikzstyleminimumheight}{8.75mm}
\newcommand{\tikzstyleminimumwidth}{12mm}

\tikzset{
    node distance=\tikzstylenodedistance,
    text centered,
    anchor=center,
}
\tikzset{
    standard node/.style n args={1}{%
        rectangle,
        rounded corners=0.1cm,
        fill=our#1,
        draw=our#1border,
        line width=0.04cm,
        minimum height=\tikzstyleminimumheight,
        minimum width=\tikzstyleminimumwidth,
        inner sep=\tikzstyleinnersep,
        text centered,
        anchor=center,
        align=center,
    }
}
\tikzset{
    standard node module/.style n args={0}{%
        rectangle,
        rounded corners=0.1cm,
        fill=ourturquoise,
        draw=ourturquoiseborder,
        line width=0.04cm,
        minimum height=\tikzstyleminimumheight, %
        minimum width=12mm, %
        inner xsep=\tikzstyleinnersep,
        inner ysep=1mm,
        text centered,
        anchor=center,
        align=center,
    }
}
\tikzset{
    standard node image/.style n args={1}{%
        rectangle,
        fill=our#1,
        draw=our#1border,
        line width=0.04cm,
        minimum height=\tikzstyleminimumheight,
        minimum width=\tikzstyleminimumwidth,
        inner sep=0,
        text centered,
        anchor=center,
        align=center,
    }
}
\tikzset{
    standard node circle/.style n args={1}{%
        fill=our#1,
        draw=our#1border,
        circle,
        inner sep=0.1cm,
        minimum height=0,
        minimum width=0,
    }
}
\tikzset{
    standard node circle/.prefix style = standard node
}

\tikzset{
    standard line/.style n args={0}{%
        line width=0.04cm,
        rounded corners=0.1cm,
    }
}
\tikzset{
    standard arrow/.style n args={0}{%
        -latex,
    }
}
\tikzset{
    standard arrow/.prefix style = standard line
}

\tikzset{
    simple node image/.style n args={0}{%
        rectangle,
        inner sep=0,
        text centered,
        anchor=center,
        align=center,
        node distance=0mm
    }
}

\let\titleold\title
\renewcommand{\title}[1]{\titleold{#1}\newcommand{\thetitle}{#1}}
\def\maketitlesupplementary
   {
    \newpage
    
    {
    \centering
    \Large
    \textbf{\thetitle}\\
    \vspace{0.5em}Supplementary Material \\
    \vspace{1.0em}
    }
   }

\begin{document}

\title{Show Me Examples:\\Inferring Visual Concepts from Image Sets} 

\titlerunning{Show Me Examples}

\author{Nick Stracke\inst{*12} \and
Kolja Bauer\inst{*12} \and
Stefan Andreas Baumann\inst{12} \and \\
Miguel Ángel Bautista\inst{3} \and
Josh Susskind\inst{3} \and
Bj\"orn Ommer\inst{12}}

\authorrunning{N.~Stracke et al.}

\institute{CompVis @ LMU \and
Munich Center for Machine Learning \and Apple}

\maketitle

\def\thefootnote{}\footnotetext{\hspace{-10pt}$^*$ Equal Contribution}
\def\thefootnote{}\footnotetext{\noindent\hspace{-10pt}All experimentation was carried out by university collaborators.}

\begin{abstract}
  Vision-language models (VLMs) can follow complex textual instructions, yet they struggle to reason from purely visual context. In particular, current models fail to infer shared concepts from sets of example images and apply them to new inputs. We introduce Visual Concept Inference from Sets (VICIS), a task that evaluates this capability. Given a small context set of images sharing a concept and a query image, the model must generate new images that preserve the context-defined concept while remaining consistent with the query. We show that state-of-the-art VLMs perform poorly on this task, often ignoring the visual context or defaulting to biased generations. To address this gap, we propose a training framework and architecture that learn to infer visual concepts from image sets and extract concept-specific embeddings from queries. Experiments on synthetic data and large-scale ImageNet/WordNet data show that our model generates more accurate and diverse outputs and generalizes to unseen concepts and modalities such as sketches.
  \keywords{Visual Reasoning \and Concept Inference \and VLMs}
\end{abstract}

\begin{figure}[h]
    \centering
    \includegraphics[width=\linewidth]{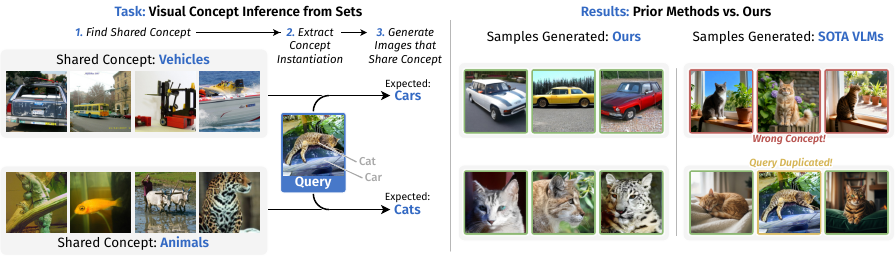}

    \caption{
    \textbf{Visual Context Inference from Sets (VICIS).}
    \textit{(left)} We introduce VICIS, a task aimed at evaluating the visual reasoning capabilities of vision models. Inspired by human capabilities, the model needs to infer a shared visual concept from a small set of example images and extract it from a query image to generate new images that preserve only this concept. \textit{(right)} Our model adapts to different contexts for the same query (e.g., \textit{vehicles} vs.\ \textit{animals}), demonstrating visual reasoning over context and query. In contrast, Nano Banana largely ignores the visual context and instead defaults to reproducing a dominant concept from the query.
    \vspace{-2em}
    }
    \label{fig:ood-example}
\end{figure}
\section{Introduction}

Generative vision-language models (VLMs) have advanced rapidly, enabling strong performance in image captioning, visual question answering, and text-conditioned image synthesis~\cite{liu2023visual, alayrac2022flamingo, chen2025janus}. A key capability underlying the success of large language models (LLMs) is in-context learning: the ability to adapt behavior at test time by conditioning on a small set of examples, without parameter updates~\cite{dong-etal-2024-survey}. If visual models are to exhibit a similar form of generalization, they must be able to infer what matters directly from visual context, even when the underlying concept is not easily verbalized.

\begin{figure}[t]
    \centering
    \includegraphics[width=0.8\linewidth]{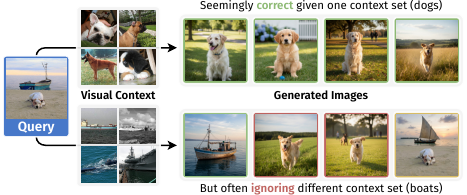}
    \caption{\textbf{Typical VLM failure cases on our VICIS task:} The model largely ignores the visual context and fixates on a single concept of the query image, in this case the dog. While the first row may suggest that it has understood the task, the second row reveals a bias toward specific concepts rather than genuine task comprehension.
    \vspace{-2em}
    }
    \label{fig:failure-case-main}
\end{figure}

Despite impressive progress, \emph{visual} in-context learning remains limited. Recent proprietary image models~\cite{batifol2025flux, fortin2025gemini, wu2025qwen} can follow many natural language instructions, yet they often fail when the instruction is specified implicitly through images (\cref{fig:failure-case-main}). In these cases, models frequently ignore the visual context, collapse to reproducing one of the inputs, or default to biased generations, suggesting that they do not reliably reason over sets of images as contextual evidence.

In this work, we isolate a concrete capability that current models struggle with, and that is central to visual reasoning: inferring a shared concept from a small set of example images and relating it to a new image. 
Humans routinely perform this type of reasoning: given a few observations, we can infer which aspects are shared and use that understanding to interpret new situations. 
Successfully solving this task requires identifying common structure across a set of images, separating relevant from irrelevant variation, and transferring the concept to a novel image.

Concretely, we introduce the task of \emph{Visual Concept Inference from Sets (VICIS)} (\cref{fig:ood-example}) to capture this form of visual reasoning. 
The model is given (i) a small \emph{context set} of images that share a common concept and (ii) a \emph{query} image. 
The goal is to generate new images that reflect the concept specified by the context while remaining consistent with the relevant aspect of the query. 
The query plays a crucial role: it provides a concrete instantiation of the concept, allowing us to assess whether the model has correctly inferred the shared structure of the context set. Without it, a model could solve the task by generating another image that resembles the context set, turning it into a task of summarizing examples rather than reasoning.%

We chose image generation as the task interface for two reasons. First, it provides a visual readout of whether the model correctly uses the context set: success requires producing outputs consistent with the context-defined concept while avoiding trivial solutions, such as copying the query. Second, it allows direct, apples-to-apples comparison to image models and VLMs that also operate in the image space, revealing a gap between general-purpose instruction following and true visual in-context reasoning (\cref{fig:failure-case-main}).

To address this gap, we design a training framework and architecture that enable us to specifically train and test this visual reasoning capability on the VICIS task. Training such a model requires constructing context sets and query-target relations that reflect shared but unlabeled concepts. We achieve this using a weak supervisory structure derived from auxiliary groupings, such as the WordNet hierarchy of ImageNet, which provides a scalable scaffold for assembling training episodes without per-concept annotation. At test time, the model operates purely on visual input: it must infer from the context set which variation is relevant and apply that inferred structure when conditioning generation on the query.

In addition to image generation, our model also predicts a compact embedding of the inferred concept. This representation provides a more direct readout of the model’s internal reasoning about the context set and allows us to analyze the structure of the inferred concept space.

We evaluate our approach on controlled synthetic data and on a large-scale hierarchical dataset constructed from ImageNet/WordNet. Across qualitative and quantitative experiments, our model generates outputs that are both more accurate and more diverse than strong baselines, and it generalizes to challenging settings such as sketches and previously unseen classes. 
These results indicate that models can be trained to use visual sets as nonverbal specifications of concepts, enabling us to express intent through examples rather than language alone.

\vspace{2mm}
\noindent We summarize our contributions as follows:%
\begin{itemize}
\vspace{-2mm}
    \item We introduce \emph{Visual Concept Inference from Sets (VICIS)}, a task that probes nonverbal visual in-context learning: inferring a concept from a small set of images and applying it to a query without text, labels, or fine-tuning.
    \item We propose a training framework that uses weak supervisory signals from auxiliary groupings to construct scalable learning episodes for this task.
    \item We present a training framework and architecture that demonstrates improved accuracy and diversity in context-guided generation, including generalization to sketches and unseen classes.
\end{itemize}

\section{Related Work}

\subsubsection{Visual In-Context Learning}
Visual in-context learning refers to the ability of visual models to adapt to new tasks using only examples provided during inference, without additional training. Although extensively studied in textual domains with large language models (LLMs)~\cite{dong-etal-2024-survey, brown2020language}, visual in-context learning remains relatively under-explored. Recent VLMs have begun addressing this challenge, yet only a handful support both unified understanding and image generation, meaning they can reason over multimodal text-image inputs and generate images. BAGEL~\cite{deng2025bagel} and ILLUME+~\cite{huang2025illume+} offer this functionality and currently represent the state of the art in unified image understanding and generation among open-source models. Very recently, the largest proprietary models have showcased surprising multi-modal capabilities. Yet, they still fail at correctly solving our proposed VICIS task, as shown in~\cref{tab:sota-vlm-eval}. 
Other approaches that explicitly target visual in-context learning, such as Visual Prompting via Image Inpainting~\cite{bar2022visual}, Improv~\cite{xu2023improv}, sequential prompting~\cite{bai2024sequential}, and Hummingbird~\cite{balazevic2023towards}, adapt via visual cues or prompts but require explicit in-context task demonstrations.

Our approach differs by removing the reliance on explicit in-context examples. Instead, our model implicitly discovers relevant concepts through provided image sets, closely resembling how humans learn from contextual examples. This framing provides a more natural and flexible mechanism for visually specifying complex, subtle, or otherwise difficult-to-describe concepts, significantly extending the scope and practicality of visual in-context learning.

\subsubsection{Inferring Latent Factors and Representations}
A large body of work studies how to uncover latent factors of variation that explain visual data at train time, often aiming for representations in which different dimensions correspond to semantically meaningful properties such as pose, lighting, or identity. 
Unsupervised latent variable models such as $\beta$-VAE~\cite{higgins2017betavae} and InfoGAN~\cite{chen2018isolating} introduce inductive biases, e.g., KL regularization or mutual-information maximization, to encourage interpretable latent structure, but are known to be non-identifiable in general~\cite{locatello2019challenging,khemakhem2020variational}. 
More recent work addresses this limitation by establishing conditions under which latent factors become identifiable, for example, by leveraging auxiliary observed variables such as iVAE~\cite{khemakhem2020variational}, or by studying related assumptions in causal representation learning~\cite{yang2020causalvae}.
In contrast to these approaches, the goal of solving the VICIS task is not to recover a universal factorization of images or identifiable latent sources of the full data distribution. 
Instead, the relevant concept needs to be inferred from a small context set of images at inference time, with the goal of extracting it from the query image.

\subsubsection{Controllable generation and concept editing.}
Another related line of work focuses on controlling generative models along semantic directions. They analyze the latent space of a pretrained generator and identify directions that induce consistent semantic changes in the output~\cite{karras2019style, goetschalckx2019ganalyze}.
Analyses of GAN latent spaces have shown that simple traversals or learned directions can induce meaningful changes in generated images~\cite{karras2019style,goetschalckx2019ganalyze}.
In diffusion models, recent work similarly finds semantic directions for more controlled generation by leveraging the conditioning mechanism~\cite{Baumann_2025_CVPR}, or by learning concept-specific adapters for precise control~\cite{gandikota2024concept,gandikota2025sliderspace}. 
These methods typically assume that the target concept is specified explicitly (e.g., via text, curated example pairs, or editing objectives) and often require additional optimization or training to enable control for a given concept~\cite{Baumann_2025_CVPR,gandikota2024concept,gandikota2025sliderspace}. 
In contrast, VICIS studies the complementary problem of \emph{inferring} the concept itself from a small set of visual examples without language or labels, and transferring it to a new query instance, without additional concept-specific training at inference time.

\subsubsection{Measuring Visual Intelligence}
Recent benchmarks like the Abstract Reasoning Corpus (ARC)~\cite{chollet2019arc}, CLEVR~\cite{johnson2017clevr}, and physics-based reasoning tests~\cite{motamed2025physicsiq} assess model intelligence by evaluating reasoning abilities on visually grounded challenges. However, these approaches typically focus on explicit input-output mappings within well-defined constraints. Our proposed VICIS task is uniquely flexible, simulating human cognitive tests by implicitly defining visual concepts through examples. It evaluates a model’s capability to generalize from abstract concept definitions to specific instantiations without explicit input-output pairs, aligning more closely with human visual intelligence assessment.

\section{Method}
Our goal is two-fold. We first want to assess the visual reasoning capabilities of current vision models. Concretely, we design a task called \textit{Visual Concept Inference from Sets} (VICIS), where the model learns to identify shared visual concepts within image sets, as outlined in~\cref{method:task_definition}. Secondly, we propose an architecture to directly solve the VICIS task, thereby providing an intuitive and effective mechanism for explicitly instructing a visual model about concepts of interest in~\cref{sec:architecture}.

\subsection{Task Definition: Visual Concept Inference from Sets}
\label{method:task_definition}

We frame our learning problem as a visual in-context task, where a model observes a small set of related images and is expected to infer the underlying visual concept they share. This setup simulates a natural form of visual learning, in which a user conveys a concept by providing example images, without relying on language or labels.

Formally, let an image \(\mathbf{x} \in \mathbf{X}\) be associated with a set of visual concepts \(C_\mathbf{x} \subseteq C\), where \(C = \{c_1, c_2, \dots, c_n\}\) denotes the complete space of possible visual concepts that can be present in an image. Each concept \(c_i\) can have multiple possible instantiations, such as \textit{car type: sedan}; \textit{surface finish: metallic}; or \textit{color: blue}.
 
To define our VICIS task, we construct a \textit{context set} \(\mathbf{X}_{\mathrm{set}}\) consisting of several images that share a common (but unlabeled) concept. For example, the context set might contain images of various dog breeds, objects in bright lighting, or items made of wood. This shared structure encourages the model to infer a latent concept that captures the variation across the set. The goal is for the model to identify and internalize the relevant visual dimension that unifies these examples.
 
In order to verify whether the model truly understands and precisely identifies the intended concept, we introduce two images: a query image \(\mathbf{x}^\mathrm{query}\) and a target image \(\mathbf{x}^\mathrm{target}\). Both the query and target images share the same specific instantiation of the concept (e.g., the same breed of dog). The role of the query image is to specify an instantiation of the relevant concept defined by the set, as illustrated in~\cref{fig:hierarchy}. The target image then acts as the correct match or answer, sharing this exact instantiation.

To enable learning of this behavior, we require a mechanism for constructing context sets and corresponding query-target pairs that reflect shared concepts. This relies on auxiliary information that enables us to group images based on shared semantics. While this introduces a form of indirect supervision, the key idea is to teach the model how to infer structure from sets of images alone. At inference time, the model is presented with only a context set and a query image, and must reason about the relevant visual concept based purely on visual input.

\begin{figure}[t]
    \centering
    \includegraphics[width=.7\linewidth]{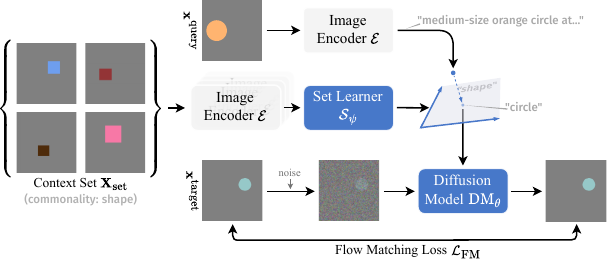}
    \caption{\textbf{Model Overview.} Given a set of images that share a concept (here, \textit{shape}), the \setlearner{} predicts a concept space that captures all possible instantiations of that visual concept. We project the embedded query image in that concept space to obtain its concrete instantiation (here, \textit{circle}) and remove all other concepts that cannot be represented in that space. Finally, the diffusion model is conditioned on the projected query to denoise the target image. 
    \vspace{-2em}
    }
    \label{fig:method}
\end{figure}
\subsection{Solving the VICIS Task}
\label{sec:architecture}

Our proposed method to solve the VICIS task comprises three components: (1) the \textit{Set Learner}, (2) the \textit{Instantiation Stage}, and (3) the \textit{Diffusion Model}. See \cref{fig:method} for an illustration. Each component plays a specialized role in our framework to discover, encode, and utilize meaningful visual concepts.

\subsubsection{Set Learner} The Set Learner identifies concept-specific directions in the feature space. Each image in the input set \(\mathbf{X}_{set}\) is first encoded individually using a pretrained Vision Transformer $\mathcal{E}$~\cite{caron2021dinov1, oquab2023dinov2, radford2021clip}, producing feature token embeddings. The Set Learner itself is a separate Vision Transformer (ViT) that takes the combined embeddings of the entire set of images as input. It reasons across these embeddings jointly to identify the shared concept and subsequently predicts a set of direction vectors \(\mathbf{D}_c = \{\mathbf{d}_1, \mathbf{d}_2, \ldots, \mathbf{d}_k\}\). These directions define the embedding space corresponding to the shared concept identified within the set.

\begin{wrapfigure}{r}{.2\linewidth}
    \vspace{-2.2em}
    \includegraphics[width=\linewidth]{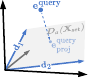}
    \caption{\todo{Instantiation Illustration}.}
    \vspace{-1em}
\end{wrapfigure}
\subsubsection{Instantiation Module} Given a query image \(\mathbf{x}^\mathrm{query}\), we encode it individually using the same pretrained encoder $\mathcal{E}$, but utilize only the CLS token embedding \(\mathbf{e}^\mathrm{query}\). 
To isolate the concept-specific information identified by the Set Learner, we project this embedding onto the embedding space spanned by the concept directions $\mathbf{d}_i$ through dot products:
\begin{equation}
    s_i = \langle \mathbf{e}^\mathrm{query}, \mathbf{d}_i \rangle, \quad \forall \mathbf{d}_i \in \mathbf{D}_c
\end{equation}
These scalars \(\mathbf{s} = \{s_1, s_2, \ldots, s_k\}\) represent the query's concept instantiation within the broader concept space. Multiplying each scalar \(s_i\) by the respective direction \(\mathbf{d}_i\) and aggregating gives a single concept-conditioned token:
\begin{equation}
    \mathbf{e}^\mathrm{query}_{proj} = \sum_{i=1}^{k} s_i \mathbf{d}_i
\end{equation}
This resulting token \(\mathbf{e}^\mathrm{query}_{proj}\) explicitly captures only the concept-specific information from the query, discarding unrelated visual details (e.g., background or unrelated objects).

\subsubsection{Diffusion Model} The diffusion model $\text{DM}_\theta$ is conditioned on \(\mathbf{e}^\mathrm{query}_{proj}\) to generate target images \(\mathbf{x}^\mathrm{target}\). Specifically, \(\mathbf{e}^\mathrm{query}_{proj}\) is added directly to the timestep embedding of the diffusion model, ensuring strong concept-conditioned generation.

To train the entire setup, we use the rectified flow objective ~\cite{lipman2023flow, albergo2023stochastic, liu2023flow}, which frames generative modeling as matching a continuous vector field. Specifically, given a data distribution $p(\mathbf{x})$ and an initial noise distribution, the diffusion model $\text{DM}$ learns a time-dependent vector field $\mathbf{v}(\mathbf{x}, t)$ that transports samples smoothly from noise toward the data manifold. Formally, this involves minimizing the Flow Matching loss:
\begin{align}
    \mathcal{L}_{\text{FM}}(\theta) &= \mathbb{E}_{t \sim \mathcal{U}[0,1],\, \mathbf{x}_{t}^\mathrm{target} \sim p_t, \mathbf{e}^\mathrm{query}_{proj}}\\\nonumber
    &\quad\left[\left\| \text{DM}_\theta(\mathbf{x}_{t}^\mathrm{target}, t, \mathbf{e}^\mathrm{query}_{proj}) - \mathbf{u}(\mathbf{x}_{t}^\mathrm{target}, t) \right\|_2^2\right]
\end{align}
This end-to-end training with a flow matching loss provides a stable objective that is easy to optimize and can work well with smaller batch sizes, which sets it apart from other methods such as contrastive learning that typically requires larger batch sizes to work successfully.

\section{Experiments}

We implement the Visual Concept Inference from Sets (VICIS) task for both controlled synthetic data and real-world data. We benchmark our model against both task-specific and general-purpose baselines, including the Visual Prompting model~\cite{bar2022visual} and recent vision-language models~\cite{huang2025illume+, fortin2025gemini, deng2025bagel}.
Our goal is to assess whether models can infer a shared visual concept from a context set and correctly extract its instantiation from a query image to generate appropriate targets.
We first study the task in a synthetic setting where the data distribution is fully controlled (\cref{subsec:vicis_toy}).
We then scale the task construction to real-world data. To that end, we propose a generic scheme to assemble context sets and query-target pairs based on auxiliary groupings from the WordNet hierarchy~\cite{wordnet}.
Finally, we analyze the concept-conditional embeddings learned by our model to gain insights into the internal representations used to solve VICIS (\cref{attribute_spaces}). Implementation Details in~\cref{app:our_imp_details}.

\begin{figure}[t]
    \centering
    \includegraphics[width=\linewidth]{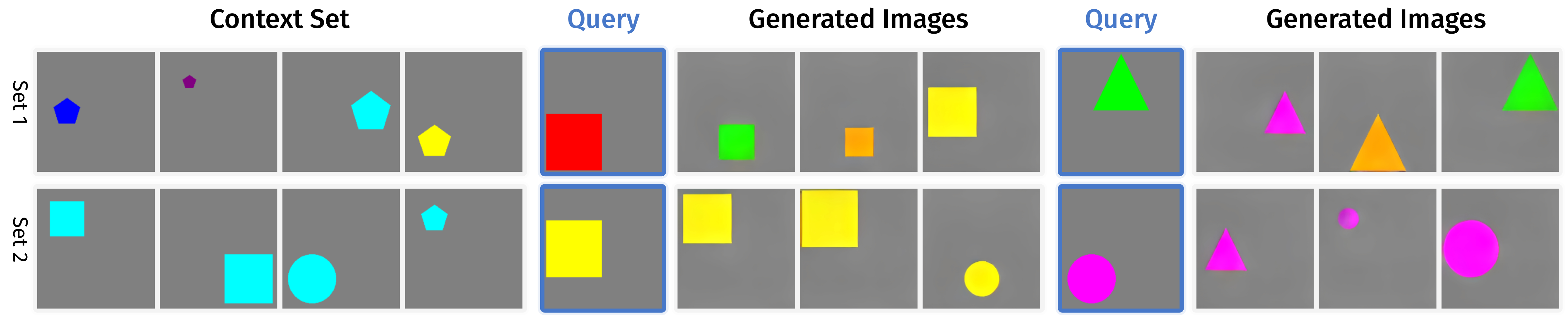}
    \caption{\textbf{Visualization of our toy dataset and results.} The left third shows the context sets used for specifying a concept, here \textit{shape} (first row) and \textit{color} (second row). The middle and right part show three samples for two different query images (left column), respectively. The model learns to correctly identify the concept specified by the context set and extract its instantiation from the query image. Note that only the specified concept's instantiation is preserved while other concepts can vary freely, indicating that the model correctly disregards irrelevant information.
    \vspace{-2em}
    }
    \label{fig:toy_example}
\end{figure}

\subsection{VICIS on Synthetic Data}
\label{subsec:vicis_toy}

We first study the VICIS task in a synthetic setting that provides full control over the visual concepts present in each image. Recent advances in generative models have made large-scale generation of high-quality synthetic visual data increasingly feasible~\cite{batifol2025flux, fortin2025gemini, yang2025qwen3}. 
In this work, we construct a simple synthetic dataset designed to isolate the core challenge of the VICIS task. This controlled environment allows us to precisely define shared concepts within context sets while varying irrelevant concepts. We illustrate the construction of such a VICIS task in the following paragraph and visualize the setup in~\cref{fig:toy_example}.

\subsubsection{Dataset Construction}
\label{data_synthetic}
To assess whether our model is generally able to reason about sets of images and find the shared concept in a context set, we construct a synthetic dataset that allows us to precisely obtain images with desired concepts for the context set and the query-target pair. Each image consists of a single object with concepts \(C_\mathbf{x} = \{\text{shape}, \text{size}, \text{location}, \text{color} \}\). We construct context set, query and target such that one concept's instantiation is shared within the context set. Query and target share a potentially different instantiation of that concept (see \cref{fig:toy_example}).

\subsection{VICIS on Real-World Data}
\label{subsec:vicis_real_world}

While the synthetic setting allows precise control over visual concepts, constructing VICIS tasks for real-world images is significantly more challenging. To scale the task beyond controlled environments, we propose a scheme for assembling context sets and query-target pairs using weakly labeled semantic structure derived from WordNet~\cite{wordnet}.

\subsubsection{Dataset Construction}
To construct meaningful concepts, we employ a hierarchical structure derived from WordNet~\cite{wordnet}. Each node in this hierarchy corresponds to a visual concept (e.g., \textit{animal}), with branches representing increasingly specific concept instantiations (e.g., \textit{mammal}$\rightarrow$\textit{dog}$\rightarrow$\textit{bulldog}). Sets are constructed at different hierarchy levels to define varying degrees of specificity for concept instantiation. For instance, if the concept \textit{animal} is chosen, a set might contain mammals, birds, and fish. Given a query image depicting a mammal, the diffusion model should generate other mammals, since that is the next level in the hierarchy. Conversely, selecting a more specific concept (e.g., \textit{dog}) restricts generation to matching exact breeds. This hierarchical approach addresses a fundamental challenge in concept inference: constructing training sets with meaningful relationships between context set, query, and target. While extensive annotations for concept instantiations would be ideal, such datasets are rare and challenging to create. Our hierarchical structure provides a flexible yet structured framework for visual concept inference, where the precise definition of sets directly shapes the model's learned capabilities. During training, context sets, queries and targets are constructed from the ImageNet training split, during evaluation they are assembled from the validation split.

\begin{figure}[t]
    \centering
    \includegraphics[width=\linewidth]{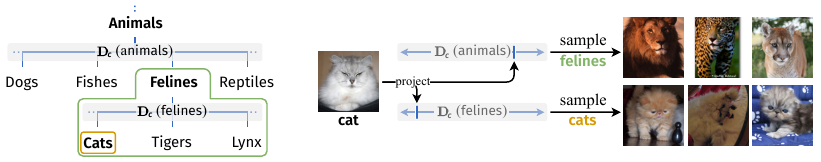}
    \caption{\textbf{Hierarchical Concept Spaces.} A simplified visualization of our ImageNet hierarchy, which is loosely based on the WordNet hierarchy. We define a concept space with respect to the next hierarchy level. The $\mathbf{D}_c(\text{animals})$ concept space differentiates roughly between different types of animals whereas $\mathbf{D}_c(\text{felines})$ is more granular and operates on a lower level of the hierarchy.
    \vspace{-1em}
    }
    \label{fig:hierarchy}
\end{figure}

\subsubsection{Quantitative Evaluation}
\label{quantitative_evaluation}
We analyze our model's performance on the VICIS task and benchmark it against both task-specific and general-purpose VLMs.
If the model correctly infers the relevant concept from the context set, it should generate targets that preserve the corresponding concept instantiation from the query. Thus, we analyze the images generated by our model to determine if the Set Learner has correctly inferred the relevant concept. In the hierarchical setting, a correct solution entails understanding the hierarchy level and branch from the context set, identifying the query’s position within the hierarchy, and generating samples from the appropriate branch. 
We compare our model against two state-of-the-art open-source VLMs,~\cite{deng2025bagel, huang2025illume+}. Since these general-purpose models have to understand the task in-context, we design three different prompting schemes shown in~\cref{tab:prompting_strategies} and ablate their performance in~\cref{tab:vlm_comparison}. The VLM results in~\cref{tab:quantitative_comparison} use the best performing prompting scheme.
Additionally, we compare against a Visual Prompting model~\cite{bar2022visual} which we specifically train for our task, as it is the closest existing method to our approach. We train the Visual Prompting model from scratch on our proposed hierarchical dataset to ensure a fair comparison. Implementation details on how we adapt their method to the proposed VICIS task are provided in~\cref{app:sec:vis_prompt}. 
We provide all models with four context images.
To quantify performance, we focus on the \textit{animal} subtree of our hierarchy. We choose this subtree for its intuitive structure, as it follows the biological taxonomy, providing a clear hierarchy for evaluation. To quantify both the correctness and diversity of generated images,~\cref{tab:quantitative_comparison} reports results for both \textit{accuracy} and \textit{diversity}:

\noindent \underline{Accuracy:} We evaluate the model’s performance on the VICIS task by measuring the correctness of the generated images.
To assess whether a generated image is in the correct part of the hierarchy, we use a classifier to obtain the class label and locate the predicted class within the hierarchy tree.
We provide the mean accuracy averaged over concepts (\textit{per con.}) and averaged over possible concept instantiations in the query (\textit{per inst.}). We observe that our model significantly outperforms the Visual Prompting model as well as the VLMs on unseen validation samples from the same hierarchy used during training, as shown in the upper section of~\cref{tab:quantitative_comparison}.

\begin{table*}[tb]
    \centering
    \caption{\textbf{Quantitative comparison of models on VICIS tasks.} Correctly solving the VICIS task entails generating correct \textit{and} diverse images. We investigate the models' ability to generalize to sketches from the ImageNet Sketch dataset~\cite{wang2019learning} and to unseen classes from ImageNet21k~\cite{imagenet}. We additionally visualize the results for the hierarchy dataset on the right to showcase the trade-off between accuracy and diversity that our model strikes best, significantly outperforming the pareto front.}
    \begin{minipage}[c]{0.63\textwidth}
        \vspace{-4mm} 
        \centering
        \adjustbox{width=\linewidth}{
            \begin{tabular}{l c cc c}
\toprule
    \multirow{2}{*}[-.2em]{Method} & \multirow{2}{*}[-.2em]{Dataset} & \multicolumn{2}{c}{Accuracy (\%) $\uparrow$} & \multirow{2}{*}[-.2em]{Diversity $\uparrow$}  \\
    \cmidrule(lr){3-4}
    & & per con. & per inst. &  \\
    \midrule
    {\color{ourgray}Baseline: Copy Query Image} & \multirow{5}{*}{Hierarchy} & {\color{ourgray}--}  & {\color{ourgray}--} & {\color{ourgray}0.47}  \\
    ILLUME+ 3B \cite{deng2025bagel} & 	& 37.15 &	55.00 &	0.57 \\
    BAGEL 7B-MoT \cite{huang2025illume+} & & 26.05 &	33.76 &	0.46 \\
    Visual Prompting~\cite{bar2022visual} &   & \underline{39.16}  & \underline{55.51} & \underline{0.84}  \\
    Ours &  & \textbf{55.68}  & \textbf{60.21} & \textbf{0.89}  \\
    \midrule
    
    Visual Prompting~\cite{bar2022visual} & \multirow{2}{*}{Sketch Queries} & 33.23 & 51.40 & \textbf{0.78}  \\
    Ours &  & \textbf{39.53} & \textbf{51.57} & 0.76  \\
    \midrule
    Visual Prompting~\cite{bar2022visual} & \multirow{2}{*}{Sketch Context} & 39.62 & 55.99 & \textbf{0.86}  \\
    Ours & & \textbf{50.73} & \textbf{57.98} & 0.85  \\
    \midrule
    Visual Prompting~\cite{bar2022visual} & \multirow{2}{*}{Sketch C+Q} & 33.61 & \textbf{51.96} & \textbf{0.80} \\
    Ours & & \textbf{38.40} & 50.37 & 0.75  \\
    \midrule
    Visual Prompting~\cite{bar2022visual} & \multirow{2}{*}{21k Context}  & 35.91 & 53.71 & 0.83 \\
    Ours &  & \textbf{47.62} & \textbf{56.53} & \textbf{0.85}  \\
    \bottomrule
\end{tabular}
        }
        \label{tab:quantitative_comparison}
    \end{minipage}%
    \hfill
    \begin{minipage}[c]{0.3\textwidth}
        \vspace{1mm}
        \centering
        \includegraphics[width=\linewidth]{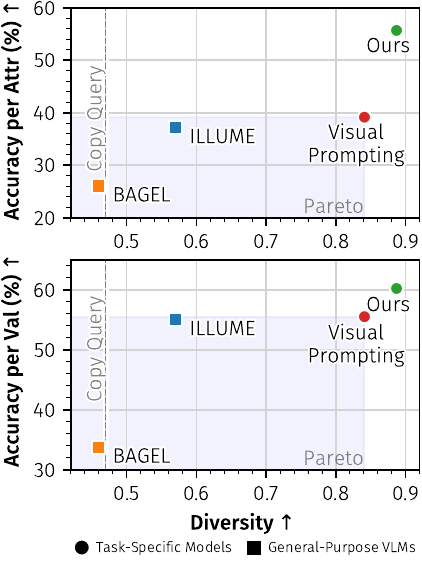} %
        \label{fig:asi_results}
    \end{minipage}
    \vspace{-5mm}
\end{table*}

\noindent \underline{Diversity:} We want the model to demonstrate understanding of the VICIS task at hand by not only producing correct but also diverse images. A trivial solution to correctly retrieve a concept's instantiation from the query is to simply retrieve all concepts' instantiations, merely copying the query. As this would not be penalized by a lower accuracy, we introduce a diversity score as a second decisive metric. This score captures the model’s ability to generate diverse but correct outputs and penalizes trivial solutions. We define the diversity score as the mean ratio of entropies between generated distributions for a single concept instantiation $q_{\text{inst}}$ projected onto two concept directions of different abstraction levels:
\begin{equation}
    \text{Div} = \frac {1} {K \cdot Q} \sum_{q_{\text{inst} \in Q}} \sum_{k=1}^{K} \frac {\text{H}(q_{\text{inst}}, \text{concept}_k^0) } {\text{H}(q_{\text{inst}}, \text{concept}_k^1) +1}
\end{equation}
$Q$ is the number of concept instantiations for the query image. $K$ denotes the number of concept pairs where $\text{concept}_k^1$ is a sub-concept of $\text{concept}_k^0$, i.e. a direct subclass in the hierarchy (e.g. $\text{concept}_k^1 = \text{fish}$ and $\text{concept}_k^0 = \text{animal}$ ). The entropy
\begin{equation}
\text{H}(q_{\text{inst}}, \text{attr}) =
-\mkern-22mu\sum_{c_i\in M_{\text{concept}}}\mkern-20mu
p(c_i \mid q_{\text{inst}}, \text{concept})
\,\log\bigl(p(c_i \mid q_{\text{inst}}, \text{concept})\bigr)
\end{equation}
is calculated over all concept instantiations $c_i \in M_{\text{concept}}$ of the higher-level concept's subtree that the concept instantiation lies in. $p(c_i | q_{\text{inst}}, \text{concept})$ denotes the probability that given a context set for concept $concept$ and a query with instantiation $q_{\text{inst}}$, the generated image's concept $concept$ is instantiated as $c_i$.  An illustration of the model's desired behaviour is provided in~\cref{fig:hierarchy}: If we project a \textit{cat} onto the concept direction \textit{feline}, we want the model to generate diverse cats. If we project that same \textit{cat} onto the concept direction \textit{animal}, we want the model to generate diverse felines, not only cats. By doing so, the model demonstrates that it has correctly inferred the context set's hierarchy level. 

To build intuition for our proposed diversity metric, we evaluate two naive baselines. Replicating the query image results in a diversity score of 0.47 and a near-perfect accuracy, only upper-bounded by the classifier’s accuracy. Generating completely random images achieves a diversity score of 0.77, comparable to the values reported in \cref{tab:quantitative_comparison}, but leads to near-zero accuracy. The diversity score's theoretical upper-bound equals 1.87 in our proposed hierarchy setup.

\subsubsection{Robustness}
\label{subsubsec:robustness}

We analyze our model’s robustness with respect to context set quality and size. When varying the number of context images from two to seven, accuracy improves steadily with additional context, particularly from two to four examples, and then plateaus. This indicates that even small sets provide useful signal, while larger sets increase reliability for more nuanced concept inference. Diversity also increases with context size. Results are shown in~\cref{tab:size}. To test sensitivity to context noise, we replace one or two images out of five with random images from the dataset. As noise increases, accuracy decreases progressively, indicating that the model effectively uses relevant context while remaining robust to partial corruption. Diversity remains stable or slightly higher, suggesting no collapse under less informative inputs. Results are shown in~\cref{tab:noise}. Overall, the model degrades gracefully under noise and benefits from larger context sets, which is consistent with the Fisher ratio trend discussed later in~\cref{fig:fisher_ratio}, suggesting that larger sets reduce ambiguity. An extended analysis of performance degradation per concept is provided in~\cref{tab:extended_robustness}.

\begin{table}[t]

\begin{minipage}[t]{0.48\linewidth}
    \centering
    \captionof{table}{Performance across context set sizes. Accuracy improves with additional context, especially in the low-data regime, and diversity increases with size.}
    \adjustbox{max width=\linewidth}{\scalebox{.85}{
        \begin{tabular}{lcccc}
        \toprule
        \multirow{2}{*}[-.2em]{\textbf{Set Size}} & \multicolumn{2}{c}{\textbf{Accuracy} (\%) $\uparrow$} & \multirow{2}{*}[-.2em]{\textbf{Div.} $\uparrow$}  \\
        \cmidrule(lr){2-3}
        & per con. & per inst. &  \\
        \midrule
        2 & 54.10 & 54.60 & 0.780 \\
        4 & 55.68 & 60.21 & 0.887 \\
        5 & 55.73 & 60.21 & 0.875 \\
        7 & 56.45 & 60.87 & 0.890 \\
        \bottomrule
    \end{tabular}
    }}
    \label{tab:size}
\end{minipage}
\hfill
\begin{minipage}[t]{0.48\linewidth}
    \centering
    \captionof{table}{Performance with noisy context sets. One or two of five context images are replaced with random images from the dataset.}
    \adjustbox{max width=\linewidth}{\scalebox{.85}{
        \begin{tabular}{lcccc}
        \toprule
        \multirow{2}{*}[-.2em]{\textbf{Set Type}} & \multicolumn{2}{c}{\textbf{Accuracy} (\%) $\uparrow$} & \multirow{2}{*}[-.2em]{\textbf{Div.} $\uparrow$}  \\
        \cmidrule(lr){2-3}
        & per con. & per inst. &  \\
        \midrule
        clean & 55.73 & 60.21 & 0.875 \\
        noisy (1/5 rand.) & 43.03 & 51.72 & 0.868 \\
        noisy (2/5 rand.) & 29.21 & 39.81 & 0.943 \\
        \bottomrule
    \end{tabular}
    }}
    \label{tab:noise}
\end{minipage}

\end{table}

\begin{figure}[t]
    \centering
    \includegraphics[width=0.8\linewidth]{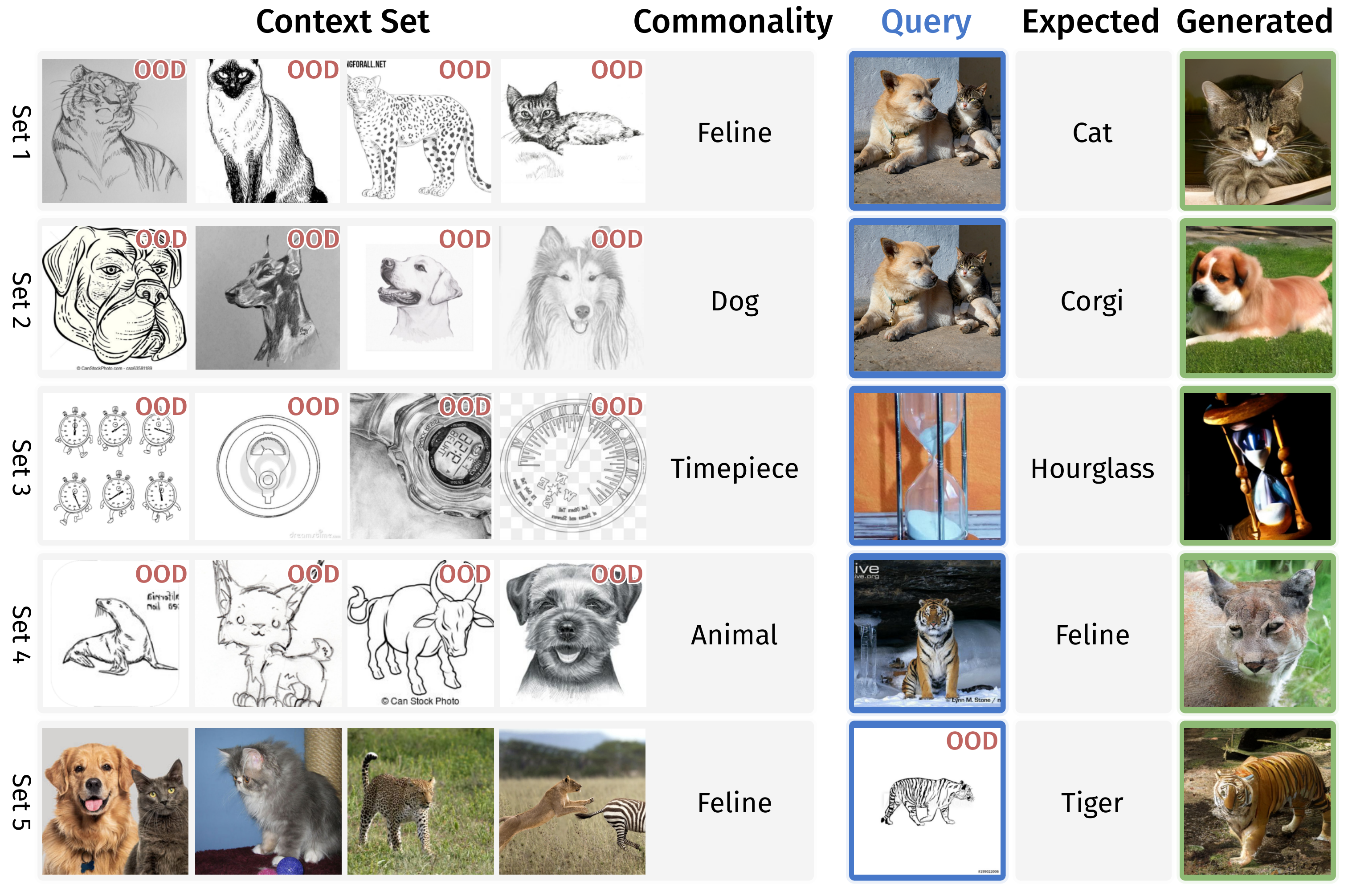}
    \caption{When a query does not fully match the concept space defined by the context set, it aligns with the closest point on the learned manifold. Although trained only on ImageNet, the model generalizes to sketch inputs by extrapolating semantically. This applies to both query and context images. In the last two rows, we show how different hierarchy levels affect generation: projecting a tiger onto the \textit{feline} space yields tiger-like samples, while projection onto the broader \textit{animal} space results in more generic felines.
    \vspace{-0.5em}
    }
    \label{fig:ood_query}
\end{figure}

\subsubsection{Generalization Capabilities}

We evaluate our model’s ability to generalize beyond its training distribution by testing it under two challenging settings: (1) using sketch images as context and/or query inputs, and (2) providing context images from previously unseen ImageNet21k classes.

To assess whether the model has overfit to appearance cues or has instead learned to capture semantic content, we use the ImageNet-Sketch dataset~\cite{wang2019learning}. We construct evaluation setups where the context set, the query, or both are drawn from sketch images. An illustration of this setup is shown in~\cref{fig:ood_query}. Quantitative results for all combinations of real and sketch images in the context and query are reported in~\cref{tab:quantitative_comparison}. While performance decreases relative to settings using only real images, our model maintains strong accuracy and consistently outperforms the Visual Prompting baseline across all configurations.

Additionally, we test if the model can correctly infer relevant concepts given novel concept instantiations not seen during training. To that end, we construct context sets from ImageNet21k and report performance in the bottom section of~\cref{tab:quantitative_comparison}. We observe only slight performance degradation, indicating that our model is able to generalize to unseen concept instantiations.

\subsection{Performance of SOTA VLMs}
\label{sec:motivation}
Since many current SOTA VLMs are closed-source and only accessible via API providers, we perform a small-scale evaluation using the FAL.ai API~\cite{falai2025} on our VICIS task. Since these models are unaware of the hierarchy, we probe them with an ambiguous query that shows two concepts and evaluate only whether they instantiate the correct concept, as specified by the context set, similar to \cref{fig:ood-example}. Since the expected output of the VICIS task is images, we can directly compare them with our model. The generated samples are evaluated and categorized by expert annotators to ensure correct labeling of all results and prevent bias from synthetic methods. We find that even current SOTA VLMs are not able to solve this task reliably, showing the need for a more principled approach (\cref{tab:sota-vlm-eval}). More details can be found in \cref{sec:app_motivational}.

\begin{table}[t]
    \centering
    \caption{\textbf{Performance of \todo{current} VLMs.} We evaluate the performance of current SOTA VLMs on a simplified version of our VICIS task and report accuracy. Even the largest closed-source VLMs are not able to solve our task reliably. Qwen and Flux do not support multiple image inputs. Tiling multiple images into one does not lead to successful task completion.}
    \resizebox{\linewidth}{!}{
    \begin{tabular}{lccccc}
        \toprule
        \textbf{Model} &
        Nano Banana\cite{fortin2025gemini} &
        Gemini 2.5 Flash Image\cite{fortin2025gemini} &
        Qwen Image Edit\cite{wu2025qwen} &
        FLUX.1 Kontext Pro\cite{batifol2025flux} &
        \textbf{Ours} \\
        \midrule
        \textbf{Accuracy} & 46\% & 40\% & not supported & not supported & \textbf{93\%} \\
        \bottomrule
    \end{tabular}
    }
    \vspace{-1em}
    \label{tab:sota-vlm-eval}
\end{table}

\subsection{VICIS for Learning Transformation}
\begin{figure}
    \centering
    \includegraphics[width=0.8\linewidth]{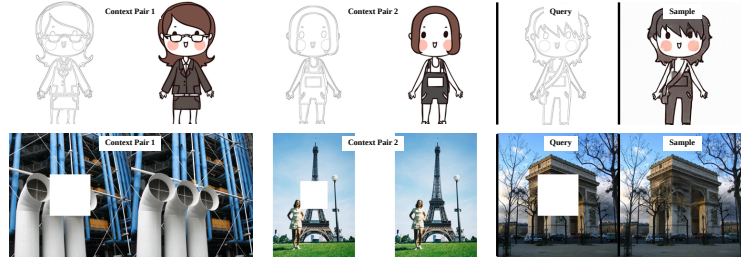}
    \caption{We show how the VICIS task can be used for more applications, such as learning transformations. Here, the context set gives multiple examples of the desired transformation, which can then be applied to the query image. We show generations for held-out query images.
    \vspace{-1em}
    }
    \label{fig:img_edit}
\end{figure}

While the previous experiments focused on concept inference, the VICIS task is a general formulation for visual reasoning. This is helpful because many useful visual instructions are easier to demonstrate than to name. In \cref{fig:img_edit}, this is shown for arbitrary image transformation pairs, i.e., visual analogy from context examples $A\rightarrow A'$ to a new query. The model is trained on Relation252k~\cite{gong2026relationadapter} with additional positional encoding for any two images that define a transformation.

\subsection{Concept-Specific Embedding Analysis}
\label{attribute_spaces}

\begin{wrapfigure}{r}{.35\linewidth}
    \vspace{-4.em}
    \centering
    \includegraphics[width=0.9718\linewidth]{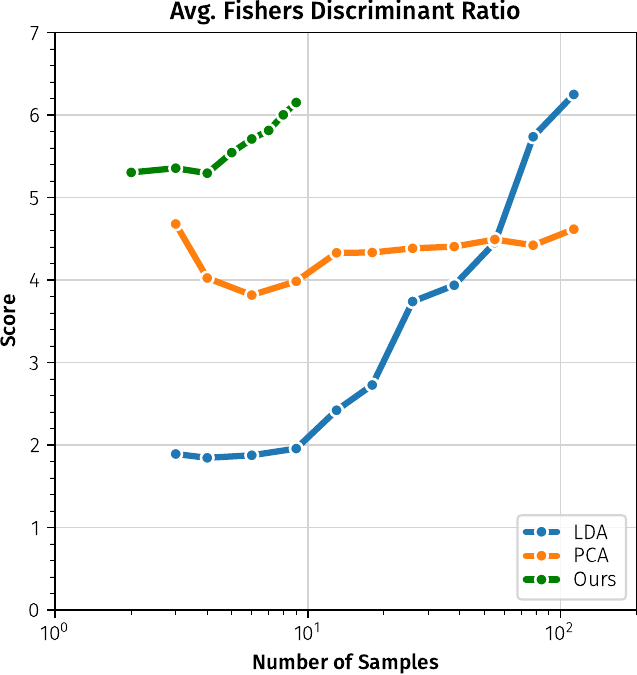}
    \caption{\textbf{Class Separation $\uparrow$} Our predicted directions better separate concepts with significantly fewer samples.}
    \vspace{-3em}
    \label{fig:fisher_ratio}
\end{wrapfigure}

Since our model predicts an embedding of the instantiated query concept, we can analyze the structure and expressiveness of that embedding space to better understand how the model solves the task. 
As described in \cref{sec:architecture}, our method predicts concept directions inferred by the Set Learner. 
This approach contrasts with traditional dimensionality reduction methods such as PCA and LDA, which derive directions from data statistics rather than explicitly learning concept-specific directions. Specifically, we evaluate how the number of examples in the context set influences performance, as larger sets should provide stronger signals for identifying the shared concept. 

To quantify the structure of the predicted concept-specific embedding space, we use the Fisher Discriminant Ratio, which assesses class separability by comparing intra-class and inter-class variance. Formally, we compute the between-class scatter matrix $\mathbf{S}_B$ and the within-class scatter matrix $\mathbf{S}_W$, where $N_c$ is the number of elements in class $c$.
We then compute the trace ratio $J_{\text{trace}}$ which provides a single scalar measure of how well-separated the concept representations are:
\vspace{2mm}
\begin{equation*}\small
    J_{\text{trace}} = \frac{\mathrm{tr}(\mathbf{S_B})}{\mathrm{tr}(\mathbf{S_W})}, \quad
    \mathbf{S}_B = \sum_{c=1}^{C}N_c(\boldsymbol{\mu}_c - \boldsymbol{\mu})(\boldsymbol{\mu}_c - \boldsymbol{\mu})^\top, \quad
    \mathbf{S}_W = \sum_{c=1}^{C}\sum_{\mathbf{x}_i \in C_c}(\mathbf{x}_i - \boldsymbol{\mu}_c)(\mathbf{x}_i - \boldsymbol{\mu}_c)^\top
\end{equation*}
\vspace{1mm}

To compare our approach with commonly used dimensionality reduction techniques, we evaluate the expressiveness of our learned concept directions against directions found by Principal Component Analysis (PCA) and Linear Discriminant Analysis (LDA). To ensure a fair comparison, we average results across multiple context sets and concept spaces. We encode images using the same pretrained encoder \(\mathcal{E}\) and linear projection layer as our method before applying PCA or LDA. While PCA captures general variance across samples, LDA explicitly optimizes for inter-class separability using class labels.

For evaluation, we use the validation split of ImageNet. We randomly sample levels from our hierarchical structure to construct context sets, which in turn define concept directions. As mentioned earlier, we average results across multiple directions to enable a fair comparison. 
We present a quantitative comparison in \cref{fig:fisher_ratio}, showing that our method produces a structured concept space comparable to LDA but requires an order of magnitude fewer samples.

\vspace{1em}
\section{Conclusion}
In this work, we introduce Visual Concept Inference from Sets (VICIS), a task designed to assess models' visual reasoning capabilities. 
Models are tasked to infer a shared concept from a small set of example images and relate it to a provided query image.
Our experiments reveal that current VLMs perform surprisingly poorly on this task, highlighting a significant gap in their ability to reason over visual examples.
To address this lack of capability, we present two schemes for constructing and training on VICIS tasks at scale using either synthetic data or weakly labeled real-world data. 
Additionally, we design an architecture that enables learning to infer these concepts from context images and extract concept-specific embeddings from query images. We demonstrate that our model successfully infers and instantiates relevant concepts and generates diverse outputs that reflect the extracted semantics. Our model outperforms both task-specific methods and general-purpose VLMs in accuracy and diversity of generated images. We validate the robustness and flexibility of our method, showing its ability to generalize to unseen concepts and modalities such as sketches.

\section*{Acknowledgments}
\todo{
This project has been supported by \todo{Apple}, the Horizon Europe project ELLIOT (GA No.\ 101214398), the project ``GeniusRobot'' (01IS24083) funded by the Federal Ministry of Research, Technology and Space (BMFTR), the BMWE ZIM-project (No.\ KK5785001LO4) ``conIDitional LoRA'', the German Federal Ministry for Economic Affairs and Energy within the project ``NXT GEN AI METHODS - Generative Methoden für Perzeption, Prädiktion und Planung'', and the bidt project KLIMA-MEMES. The authors gratefully acknowledge the Gauss Center for Supercomputing for providing compute through the NIC on JUWELS/JUPITER at JSC and the HPC resources supplied by the NHR@FAU Erlangen.

Further, we would like to thank Owen Vincent for continuous technical support.
}

\bibliographystyle{splncs04}
\bibliography{main}

\clearpage

\clearpage
\setcounter{page}{1}
\maketitlesupplementary
\setcounter{figure}{0}
\setcounter{table}{0}
\setcounter{equation}{0}
\setcounter{section}{0}
\renewcommand\thefigure{\Alph{figure}}  
\renewcommand\thetable{\Alph{table}}  
\renewcommand\thesection{\Alph{section}}  

\crefalias{section}{appsec}
\crefalias{figure}{appfig}
\crefalias{table}{apptab}
\crefalias{equation}{appeq}

\section{Multiple Shared Concepts}

Real-world scenarios often involve context sets with multiple shared concepts, creating ambiguity about which concepts to preserve. 
To evaluate our model's robustness in such a setting, we conduct controlled experiments using our synthetic datasets from Section 4.1.
We construct context sets with exactly one or exactly two shared concepts, enabling precise evaluation of the model's behavior under ambiguous conditions. The dataset contains four concepts: shape (5 possible instantiations), size (7), position (9), and color (10).

\subsubsection{Experimental Setup:} We generate 100 queries and corresponding context sets, sampling 100 output images per query-context combination (10,000 total images). We measure:
\begin{itemize}
\item \textbf{Accuracy}: Preservation of instantiations of shared concepts using pretrained classifiers ($>98\%$ validation accuracy)  
\item \textbf{Entropy}: Diversity of non-shared concepts, with bounds from query replication (0) to uniform distribution (2.01)
\end{itemize}

\begin{table}[h]
\centering
\caption{Performance evaluation on synthetic dataset with single and multiple shared concepts. Strict accuracy requires both concept instantiations to match simultaneously.}
\resizebox{\linewidth}{!}{%
\begin{tabular}{lccc}
\toprule
\textbf{Setting} & \textbf{Mean Accuracy} (\%) $\uparrow$ & \textbf{Mean Entropy} $\uparrow$ & \textbf{Random Guessing} (\%) \\
\midrule
1 shared concept & 90.96 & 1.88 & 13.85 \\
2 shared concepts (strict) & 90.57 & 1.77 & 1.87 \\
\bottomrule
\end{tabular}
}
\label{tab:multiple_concepts_synthetic}
\end{table}

For the two-concept setting, we report strict accuracy where a generated image receives 100\% accuracy only if both predicted concept instantiations match the ground truth, and 0\% otherwise. This evaluation ensures that the model genuinely preserves multiple concepts simultaneously rather than succeeding on individual concepts independently.

The model maintains high accuracy even with two shared concepts while preserving substantial diversity (entropy well above replication baseline). Notably, accuracy remains stable when transitioning from one to two shared concepts, indicating that richer conditioning signals help the diffusion model extract shared properties more precisely without sacrificing diversity. Results are presented in~\cref{tab:multiple_concepts_synthetic}.

\section{Extended Robustness Analysis}

We extend the analysis of our model's robustness to context set quality and size from~\cref{subsubsec:robustness}. To that end, we analyze performance degradation per concept in~\cref{tab:extended_robustness}.

\begin{table}[t]
\label{tab:extended_robustness}
\centering
\caption{Extended analysis of robustness to context set quality and size, measured per concept. For varying context noise and context sizes, we report the accuracy and diversity deltas to our main model per concept.}
\scriptsize

\begin{tabular}{lrrrrrrrrrr}
\toprule
\multirow{2}{*}{Context Set}
& \multicolumn{10}{c}{Accuracy (\%) $\uparrow$} \\
\cmidrule{2-11}
& arachnid & bird & dog & feline & fish & insect & primate & reptile & snake & avg. \\
\midrule
clean (\#ctx = 5)    & $81.67$ & $58.25$ & $51.65$ & $83.67$ & $62.43$ & $63.60$ & $61.62$ & $74.67$ & $61.88$ & $60.21$ \\
\midrule
noisy (1/5 rand.)    & $-5.00$ & $-9.17$ & $-5.58$ & $-2.48$ & $-7.63$ & $-18.28$ & $-14.07$ & $-9.38$ & $-9.68$ & $-8.48$ \\
noisy (2/5 rand.)    & $-17.11$ & $-27.89$ & $-11.57$ & $-18.52$ & $-20.97$ & $-23.17$ & $-26.25$ & $-37.35$ & $-22.94$ & $-20.40$ \\
\midrule
\#ctx = 3    & $-1.56$ & $-0.07$ & $-0.70$ & $-0.82$ & $+0.57$ & $-1.43$ & $+2.10$ & $-1.19$ & $-3.18$ & $-0.60$ \\
\#ctx = 2    & $-6.61$ & $-6.42$ & $-4.53$ & $-0.64$ & $-4.71$ & $-12.15$ & $-4.40$ & $-0.73$ & $-12.15$ & $-5.60$ \\
\bottomrule
\end{tabular}

\vspace{0.8em}

\begin{tabular}{lrrrrrrrrrr}
\toprule
\multirow{2}{*}{Context Set}
& \multicolumn{10}{c}{Diversity $\uparrow$} \\
\cmidrule{2-11}
& arachnid & bird & dog & feline & fish & insect & primate & reptile & snake & avg. \\
\midrule
clean (\#ctx = 5)    & $1.06$ & $0.94$ & $0.83$ & $0.98$ & $0.70$ & $0.91$ & $0.89$ & $0.86$ & $0.91$ & $0.88$ \\
\midrule
noisy (1/5 rand.)    & $\pm0.00$ & $-0.07$ & $+0.13$ & $-0.03$ & $+0.10$ & $-0.27$ & $-0.03$ & $-0.11$ & $-0.08$ & $-0.01$ \\
noisy (2/5 rand.)    & $-0.36$ & $-0.03$ & $+0.47$ & $-0.36$ & $-0.05$ & $-0.20$ & $-0.12$ & $-0.24$ & $-0.23$ & $+0.07$ \\
\midrule
\#ctx = 3    & $+0.02$ & $+0.02$ & $-0.01$ & $-0.07$ & $-0.01$ & $-0.01$ & $+0.07$ & $-0.01$ & $-0.09$ & $-0.01$ \\
\#ctx = 2    & $-0.11$ & $-0.12$ & $-0.07$ & $-0.08$ & $-0.04$ & $-0.23$ & $-0.06$ & $-0.02$ & $-0.22$ & $-0.10$ \\
\bottomrule
\end{tabular}

\end{table}

\section{General Implementation Details}
\label{app:our_imp_details}

We train the entire setup end-to-end. The set learner is parameterized as a ViT-L~\cite{dosovitskiyvit} and the diffusion model as a SiT-L~\cite{albergo2023stochastic}. In total this leads to 765M trainable parameters. We train our model on 16 H100 GPUs for \todo{24} hours using 30 query-target image pairs for every set and a global batch size of 80 sets. As the Set Learner only needs to run once per set, we precompute the concept directions for a set and reuse them for all 30 query-target image pairs. Our embedding space has four directions of dimensionality 256. The image encoder $\mathcal{E}$ is a ViT-L which we initialize with DINOv2~\cite{darcet2024vision,caron2021dinov1} for faster convergence. We use Adam~\cite{kingma2014adam} as our optimizer with a learning rate of 1e-4 with linear warmup.

\section{Additional VLM Evaluation Details}
\label{sec:app_motivational}
We use the FAl.ai API~\cite{falai2025} to access closed-source models. Due to the cost of using those APIs, we only generate 100 images per model and categorize their output as shown in \cref{tab:app_vlm_eval}. As there are no reliable classifiers for categorizing the generated samples, this is done manually by expert annotators. Since these models have no way of knowing the ImageNet hierarchy, we provide a simplified hierarchy in the prompt to make the task more comparable to our main evaluation. The prompt used is: ``I will provide you with 5 images. The first four images all contain one shared concept, i.e., they have something in common. Your goal is to identify this shared concept and find the corresponding level in the provided hierarchy. Then take a look at the fifth image and find the correct child branch of the selected hierarchy level, i.e., the correct first level under it. Then generate an image showing a random instance from that level. The hierarchy: [hierarchy]''. Since we only measure the accuracy of the model, the hierarchy is not strictly required as the model does not need to sample diverse images from the correct hierarchy level, but we found that including the hierarchy increases performance. Since the closed-source models support single-turn prompting~(\cite{fortin2025gemini}), we do not use the prompts from our ablation in~\cref{tab:prompting_strategies}.

\begin{table}[h]
    \centering
    \caption{\textbf{Error Evaluation.} We roughly classify where the evaluated VLMs fail. In most cases, the model is not able to understand the context set and find the common concept, leading it to also generate the wrong concept. In rarer cases, the model simply replicates the query image or produces an unrelated image to the query and context set.}

    \begin{tabular}{lccc}
        \toprule
        & \textbf{BAGEL}\cite{deng2025bagel} & \textbf{ILLUME+}\cite{huang2025illume+} & \textbf{Ours} \\
        \midrule
        Replicate Query Image   & 33 & 21 & 0 \\
        Sample Wrong Concept    & 10 & 18 & 0 \\
        Sample Unrelated Image  & 30 & 36 & 7 \\
        Sample Correct Concept  & 27 & 25 & 93 \\
        \% Correct              & 27\% & 25\% & \textbf{93\%} \\
        \bottomrule
    \end{tabular}
    
    \label{tab:app_vlm_eval}
\end{table}

\section{Concept-Specific Embedding Space Interpretability}
\begin{figure}[t]
    \centering
    \begin{tabular}{ccc}
    \toprule
        Concept Instantiation $a$ & Interpolation $a \xrightarrow{}b$ & Concept Instantiation $b$ \\  
    \midrule
        \includegraphics[width=0.3\linewidth]{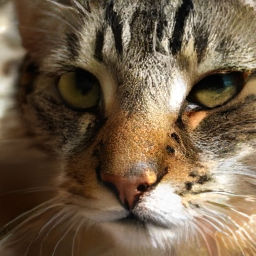} & \includegraphics[width=0.3\linewidth]{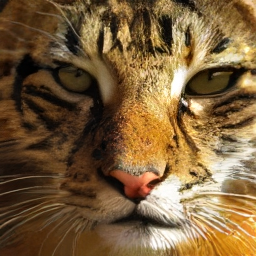} & \includegraphics[width=0.3\linewidth]{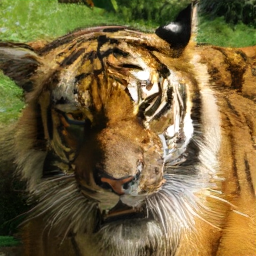} \\

        \includegraphics[width=0.3\linewidth]{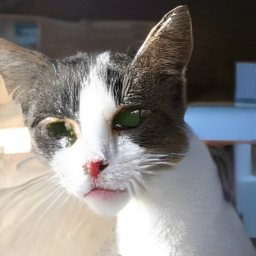} & \includegraphics[width=0.3\linewidth]{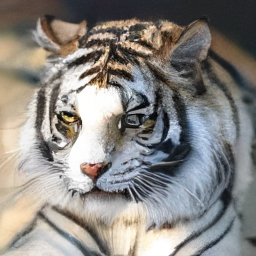} & \includegraphics[width=0.3\linewidth]{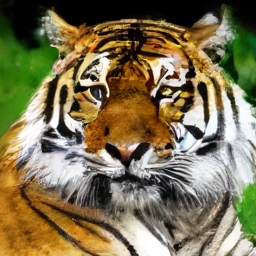} \\

    \bottomrule
    \end{tabular}
    \caption{Interpolation examples between two concept instantiations. }
    \label{fig:interpolation}
\end{figure}

We qualitatively explore how semantically meaningful the predicted concept directions are by interpolating between two concept instantiations. We first select a relevant concept to define the embedding space for the interpolation, e.g., \textit{feline}, project images of cats and tigers in the embedding space, and find a direction by taking the average delta between the embedded tiger and cat points. We then sample points from this direction and conditionally generate images using the diffusion model. In ~\cref{fig:interpolation} we show examples of such interpolations.

\section{Prompt Ablation Illume+}
\label{sec:vlm_comparison}

To establish the effectiveness of our approach relative to existing multimodal systems, we conduct a comprehensive evaluation against ILLUME+~\cite{huang2025illume+}, a state-of-the-art vision-language model that supports both image understanding and generation tasks.

Since ILLUME+ cannot perform both image understanding (context set analysis) and image generation in a single inference round like our method, we design a two-stage prompting protocol. In the first round, the model analyzes the context set to identify shared concepts. In the second round, it generates a new image based on the identified concept and the query image. We evaluate three distinct prompting strategies to ensure a fair comparison and follow our standard evaluation protocol described in \cref{quantitative_evaluation}.

\cref{tab:vlm_comparison} presents the quantitative results comparing different prompting strategies for ILLUME+ against our method. Our approach strikes a significantly better trade-off across evaluation metrics compared to all prompting variants. The best-performing prompting strategy (Prompt \#2) achieves 37.15\% accuracy per concept, 55.00\% accuracy per instantiation, and a diversity of 0.568, while our method achieves 46.38\%, 54.46\%, and 0.81, respectively. 
We detail the three prompting strategies employed for ILLUME+ in \cref{tab:prompting_strategies}.

\begin{table}[t]
\centering
\caption{Performance comparison of three prompting strategies for ILLUME+ versus our method. For all metrics, higher is better ($\uparrow$).}
\label{tab:vlm_comparison}
\resizebox{\linewidth}{!}{%
\begin{tabular}{lcccc}
\toprule
\textbf{Method} & \textbf{Acc. per Con. (\%) $\uparrow$} & \textbf{Acc. per Inst. (\%) $\uparrow$} & \textbf{Diversity} $\uparrow$ \\
\midrule
ILLUME+ Prompt \#1 & 33.17 & 51.08 & 0.578  \\
ILLUME+ Prompt \#2 & \underline{37.15} & \textbf{55.00} & \underline{0.568}  \\
ILLUME+ Prompt \#3 & 33.19 & 50.56 & 0.575  \\
\midrule
\textbf{Ours} & \textbf{46.38} & \underline{54.46} & \textbf{0.81} \\
\bottomrule
\end{tabular}
}
\end{table}

\begin{table}[t]
\centering
\caption{Prompting strategies used for ILLUME+ evaluation.}
\label{tab:prompting_strategies}
\resizebox{\textwidth}{!}{%
\begin{tabular}{cp{0.4\textwidth} @{\hspace{0.02\textwidth}} p{0.4\textwidth}}
\toprule
\textbf{Prompt} & \textbf{Round 1: Concept Inference} & \textbf{Round 2: Image Generation} \\
\midrule
\textbf{\#1} & 
\begin{minipage}[t]{0.4\textwidth}
\textit{Text:} ``I provided you with four images and a class hierarchy below. Find the hierarchy level that best matches the commonality / shared concept in the four images. Answer only with the name of the hierarchy level, nothing else. The hierarchy: [hierarchy]''\\
\textit{Input:} Context set images
\end{minipage} & 
\begin{minipage}[t]{0.4\textwidth}
\textit{Text:} ``I provided you with one image and a class hierarchy below. Find out what child of the [extracted\_concept] hierarchy level best matches the content of the provided image. Generate an image showing a random instance from that level. If there are no children under the selected hierarchy level, you can generate an image from the same class of the given image. \\ The hierarchy: [hierarchy]''\\
\textit{Input:} Query image
\end{minipage} \\
\midrule
\textbf{\#2} & 
\begin{minipage}[t]{0.4\textwidth}
\textit{Text:} ``Here are four images and a taxonomy. Identify the hierarchy level that captures the common concept present in all four images. Answer only with the name of the hierarchy level, nothing else. \\ The hierarchy: [hierarchy]''\\
\textit{Input:} Context set images
\end{minipage} & 
\begin{minipage}[t]{0.4\textwidth}
\textit{Text:} ``I will give you one image along with a hierarchy of categories. Identify which child node of the hierarchy level [extracted\_concept] best describes the image. After identifying it, generate a new image showing a random example from that node. If there are no children under the selected hierarchy level, you can generate an image from the same class of the given image. \\ The hierarchy: [hierarchy]''\\
\textit{Input:} Query image
\end{minipage} \\
\midrule
\textbf{\#3} & 
\begin{minipage}[t]{0.4\textwidth}
\textit{Text:} ``I am giving you four example images and a hierarchy of categories. Find the hierarchy level that represents the concept or class they all have in common. Answer only with the name of the hierarchy level, nothing else. \\ The hierarchy: [hierarchy]''\\
\textit{Input:} Context set images
\end{minipage} & 
\begin{minipage}[t]{0.4\textwidth}
\textit{Text:} ``Analyze the given image in the context of the following hierarchy. Pick the child under the hierarchy level [extracted\_concept] that best matches this image. Then produce an image showing another random instance from that same hierarchy level. If there are no children under the selected hierarchy level, you can generate an image from the same class of the given image. \\ The hierarchy: [hierarchy]''\\
\textit{Input:} Query image
\end{minipage} \\
\bottomrule
\end{tabular}%
}
\end{table}

Our results reveal several key insights. First, prompt optimization significantly impacts VLM performance, with Prompt \#2 outperforming Prompt \#3 substantially in per concept accuracy. However, even the best-performing prompt falls substantially short of our method's performance overall. Second, the two-stage nature of the VLM approach introduces potential error propagation, where mistakes in concept identification compound during image generation. Finally, our end-to-end approach demonstrates superior capability in both understanding shared concepts and generating coherent images that preserve these concepts while introducing appropriate variations.

\section{Few-shot Classification}
\label{sec:downstream_tasks}

The VICIS framework naturally extends to various downstream applications through its inference-time adaptability. We demonstrate this capability through a few-shot classification experiment. We evaluate our method's transferability by applying a model trained on the ImageNet hierarchy to few-shot classification on the Caltech-UCSD Birds (CUB) dataset~\cite{WahCUB_200_2011}. A key advantage of VICIS is its ability to adapt the encoder at inference time by defining context sets without requiring additional training or class information.

Our experimental setup uses context sets of five randomly selected bird images to define a bird-specific embedding space. We then perform 1-shot 5-way classification by projecting CUB test images in this predicted embedding space. This approach achieves an accuracy of \textbf{72.28\%}.

For comparison, we establish a DINO baseline using an equivalent experimental setup. We create a DINO space by embedding the same context images and applying PCA with four components (matching our number of concept directions). Few-shot classification in this DINO space yields a baseline accuracy of \textbf{71.78\%}. Our method outperforms this strong baseline despite being trained on ImageNet at a much smaller scale than DINO, demonstrating the effectiveness of our concept-specific embeddings.

\section{Additional Qualitative Examples}
\begin{figure}[b]
    \centering
    \includegraphics[width=0.55\linewidth]{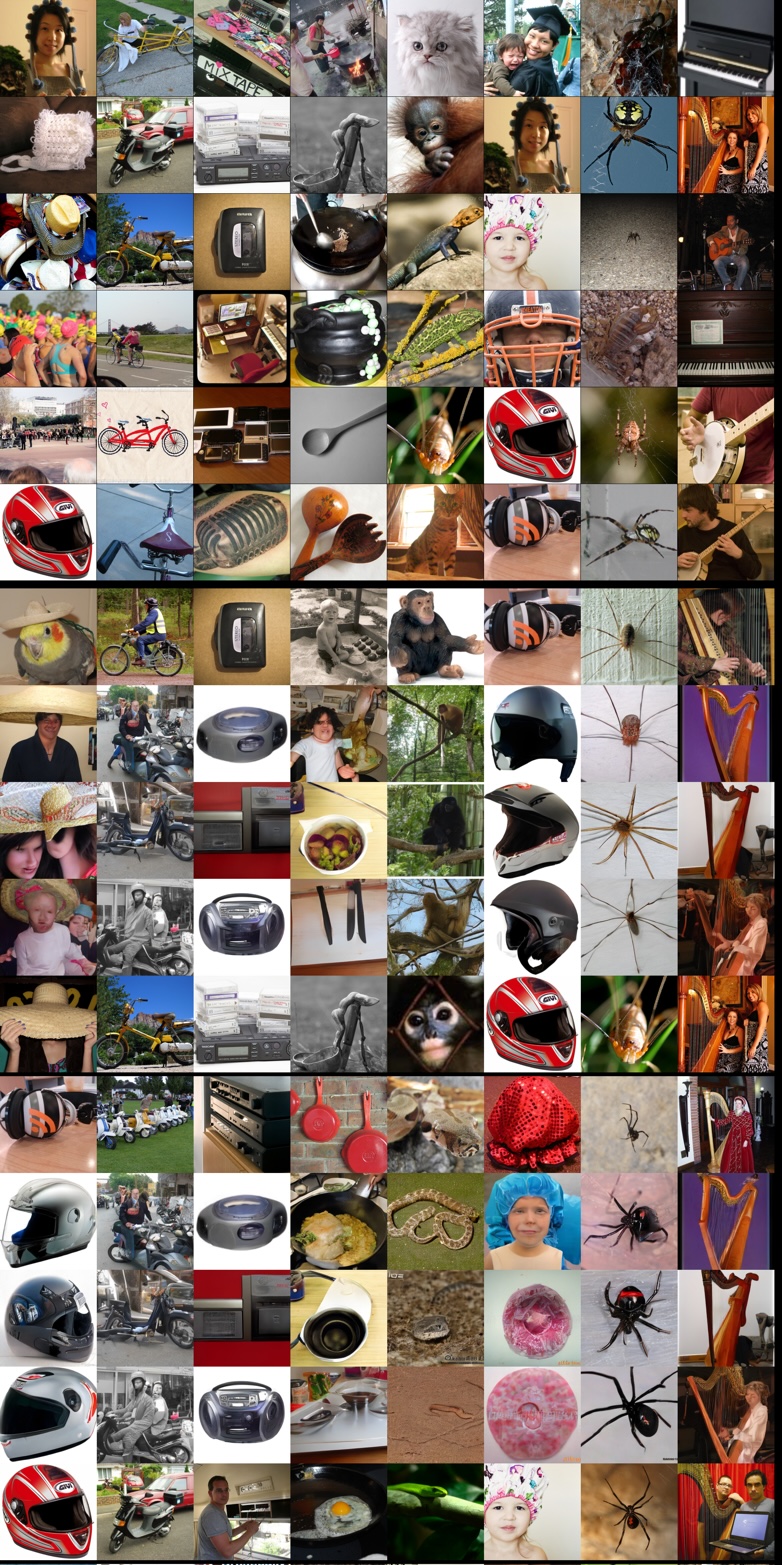}
    \caption{Generated samples from our model alongside corresponding context sets and query images. The upper third shows the context sets (column-wise) consisting of six images each. The following two sections are organized in the following way: query, three samples, and target. This visualization intends to provide a better intuition of how the hierarchy is structured and how we compose sets, queries, and targets.}
    \label{fig:ours_reel}
\end{figure}

We show additional qualitative examples from our hierarchical Imagenet model in \cref{fig:ours_reel}.

\section{Visual Prompting Implementation Details}
\label{app:sec:vis_prompt}

To ensure compatibility with~\cite{bar2022visual}, we cast the VICIS task as a 3x2 grid image that encompasses a 2x2 context set at the top, a query image on the bottom left and the denoising target image on the bottom right. We train a flow matching model to denoise the bottom right part of this 3x2 grid. To enable a fair comparison to our main model, the flow matching model is parameterized as a SiT-L~\cite{albergo2023stochastic}. Similar to our main model, the context and query images from the 3x2 grid are individually encoded by a ViT-L image encoder $\mathcal{E}$ which we initialize with DINOv2~\cite{darcet2024vision,caron2021dinov1} for faster convergence. We train the model on 16 H100 GPUs for 24 hours.

\clearpage

\end{document}